\documentclass{article}

\usepackage{microtype}
\usepackage{graphicx}
\usepackage{subfigure}
\usepackage{booktabs} 

\usepackage{hyperref}
\usepackage{dot2texi}
\usepackage{tikz}

\usepackage[accepted]{icml2025}

\usepackage{amsmath}
\usepackage{amssymb}
\usepackage{mathtools}
\usepackage{amsthm}

\usepackage[capitalize,noabbrev]{cleveref}
\usepackage{xspace}
\usepackage{colortbl}
\definecolor{Gray}{gray}{0.9}

\usepackage{xurl}

\theoremstyle{plain}
\newtheorem{theorem}{Theorem}[section]
\newtheorem{proposition}[theorem]{Proposition}
\newtheorem{lemma}[theorem]{Lemma}
\newtheorem{corollary}[theorem]{Corollary}
\theoremstyle{definition}
\newtheorem{definition}[theorem]{Definition}
\newtheorem{assumption}[theorem]{Assumption}
\theoremstyle{remark}
\newtheorem{remark}[theorem]{Remark}

\usepackage[textsize=tiny]{todonotes}

\icmltitlerunning{Under review in ICML 2025}

\begin{document}

\twocolumn[
\icmltitle{Pay Attention to What You Need}



\icmlsetsymbol{equal}{*}

\begin{icmlauthorlist}
\icmlauthor{Yifei Gao}{equal}
\icmlauthor{Shaohong Chen}{equal}
\icmlauthor{Lei Wang$\dag$}{}
\icmlauthor{Ruiting Dai}{}
\icmlauthor{Ziyun Zhang}{}
\icmlauthor{Kerui Ren}{}
\icmlauthor{Jiaji Wu}{}
\icmlauthor{Jun Cheng}{}
\end{icmlauthorlist}



\icmlkeywords{Machine Learning, ICML}

\vskip 0.3in
]




\begin{abstract}
Although large language models (LLMs) have achieved significant success in natural language processing, they still struggle with long-context comprehension. Traditional approaches to mitigating this issue typically rely on fine-tuning or retraining, which is both resource-intensive and challenging to deploy in lightweight industrial settings. In this paper, we investigate the potential to accomplish this without any additional resources. Through an in-depth study of the attention mechanism in LLMs, we propose a method called \textbf{S}caled \textbf{R}e\textbf{A}ttention (SRA) to strengthen LLMs' ability to interpret and retrieve information by strategically manipulating their attention scores during inference. Through extensive experiments, we demonstrate that integrating SRA significantly boosts LLMs’ performance on a variety of downstream tasks, highlighting its practical potential for enhancing language understanding without incurring the overhead of traditional training.
\end{abstract}

\section{Introduction}
Large language models (LLMs) with attention mechanisms~\cite{openaiannouncement} have achieved tremendous success across a wide range of downstream tasks in recent years. Their success can largely be attributed to the superiority of the attention architecture~\cite{vaswani2017attention}. However, as tasks become more complex and the required contextual understanding increases, LLMs often fall short. 

When the input length exceeds a certain limit, LLMs often ``forget" previously mentioned content or experience ``memory confusion," leading to incorrect outputs. Even with prompt engineering techniques like Chain of Thought (CoT)~\cite{nye2021show,wei2022chain}, the models still struggle with complex problems. This limitation originates inherently from the model itself, making it unavoidable through fine-tuning or retraining—both of which demand substantial resources. This inspired the motivation for this paper: \textit{enhancing the model's comprehension and retrieval capabilities without additional training.}

\begin{figure*}[t]
    \centering
     \includegraphics[width=0.9\linewidth]{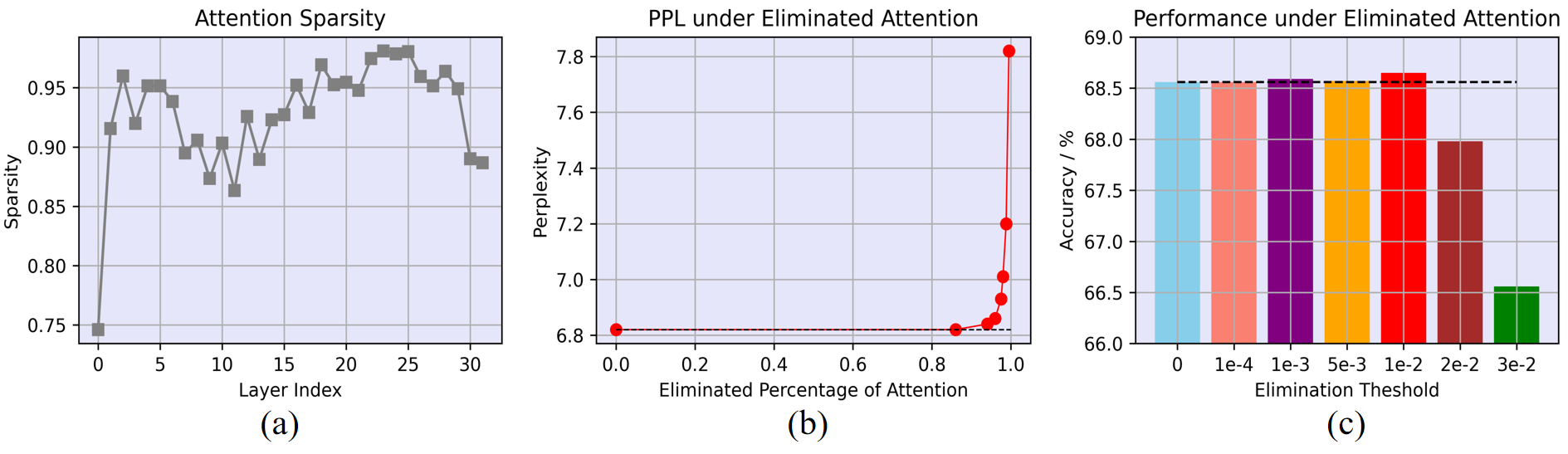}
    \caption{Characteristics of attention in LLaMA-3-8B: (a) The sparsity level of attention in each layer, with the sparsity threshold set at 0.001 on text length 2048. (b) The perplexity of WikiText2 on text length 2048 after attention elimination. (c) Averaged performance on downstream tasks (ARC, PIQA, Hellaswag, Winogrande) after attention elimination. Even with 25$\%$ of attention weights eliminated (threshold 2e-2), the performance remains nearly unchanged. The black dashed line represents the original performance.}
    \label{fig:attn_sparse_elimi}
\end{figure*}

We began by identifying the attention mechanism as the critical component for retrieving and interpreting context within LLMs. Building on our empirical findings and existing research \cite{wang2020linformer,zandieh2023kdeformer}, we noted that most tokens—and their corresponding attention scores—have a negligible effect on the model’s reasoning. Even after eliminating the majority of these scores, the model’s performance remained nearly unchanged, as illustrated in Figure~\ref{fig:attn_sparse_elimi}. Intuitively, if we can better utilize the ``wasted" attention scores, the model should achieve improved performance. By manually adjusting attention scores during inference and accepting a slight trade-off in model stability, we achieved a significant improvement in comprehension and retrieval capabilities, all without any fine-tuning, retraining, or auxiliary resources. To the best of our knowledge, \textbf{this represents the first effort to address these challenges from such a perspective}.

In this paper, we introduce \textbf{Scaled ReAttention} (SRA), a technique that first discards unimportant attention scores and then redirects them toward more informative tokens. During this process, SRA strategically relaxes the model’s inherent stability, leveraging the elimination results to further enhance its comprehension. Our technique is plug-and-play and could be integrated into a wide range of existing LLMs. With SRA, we successfully improved the performance of LongChat-7B-16K and LLaMA-3-8B on the LongChat retrieval task by over 10\% compared to the original models. Additionally, we significantly outperformed the original models with LLaMA-3-8B-Instruct and LLaMA-2-13B-Chat on the XSUM summarization task. Furthermore, on the public datasets such as LongBench v1 (v2), we improved the performance of a series of LLMs by above 1.5\%.

Our contribution can be concluded as follows:
\vspace{-10pt}
\begin{itemize}
    \setlength{\itemsep}{-3pt}
    \item A comprehensive analysis of the attention mechanism and attention scores in LLMs, offering foundational insights into the SRA technique.
    \item A novel plug-and-play method that enhances the comprehension and retrieval capabilities of LLMs without the need for fine-tuning or retraining.
    \item Empirical evidence from extensive experiments showcasing SRA's ability to significantly improve performance in a variety of tasks.
\end{itemize}

\section{Related Work}

\subsection{Strengthen Long-Context Comprehension}
Prior work has primarily shown how better training methods~\cite{zhang2021deep,wang2022self} or larger datasets~\cite{hoffmann2022training,gpt4} can be used to improve model performance. Despite promising results, their excessive reliance on human and computational resources imposes significant limitations on their industrial applications.

On the other hand, solving relevant issues by retrieval~\cite{izacard2023atlas,jiang2022retrieval} to locate the main content while discarding irrelevant information can be equally effective. However, these approaches often require additional training of a ``retriever"~\cite{karpukhin2020dense} to assist with retrieval and are powerless when addressing problems that demand improved model understanding.

\subsection{Extend Context Window}
Previous research has highlighted the critical role of positional encoding (PE) in model performance~\cite{vaswani2017attention,su2023roformer,ni2021t5}, as PE conveys essential information about the relationships between tokens. However, this adaptability can introduce substantial disruption when handling text that exceeds the model’s pretraining length~\cite{press2021alibi}. To address this, methods such as Position Interpolation (PI)~\cite{chen2023pi,emozillareddit} have been proposed to extend RoPE by creating intermediate angles. Meanwhile, LandMark Attention\cite{mohtashami2023landmark} incorporates an additional “Landmark” token for block-wise information representation, which slightly modifies the underlying model structure.

Although these approaches effectively broaden the context window of LLMs without introducing extensive additional resources, their achievements are at the expense of the model's performance on downstream tasks, which severely limits their practical applications. However, with the method proposed in this paper, their performance can be substantially improved.

\begin{figure}
    \centering
     \includegraphics[width=0.8\linewidth]{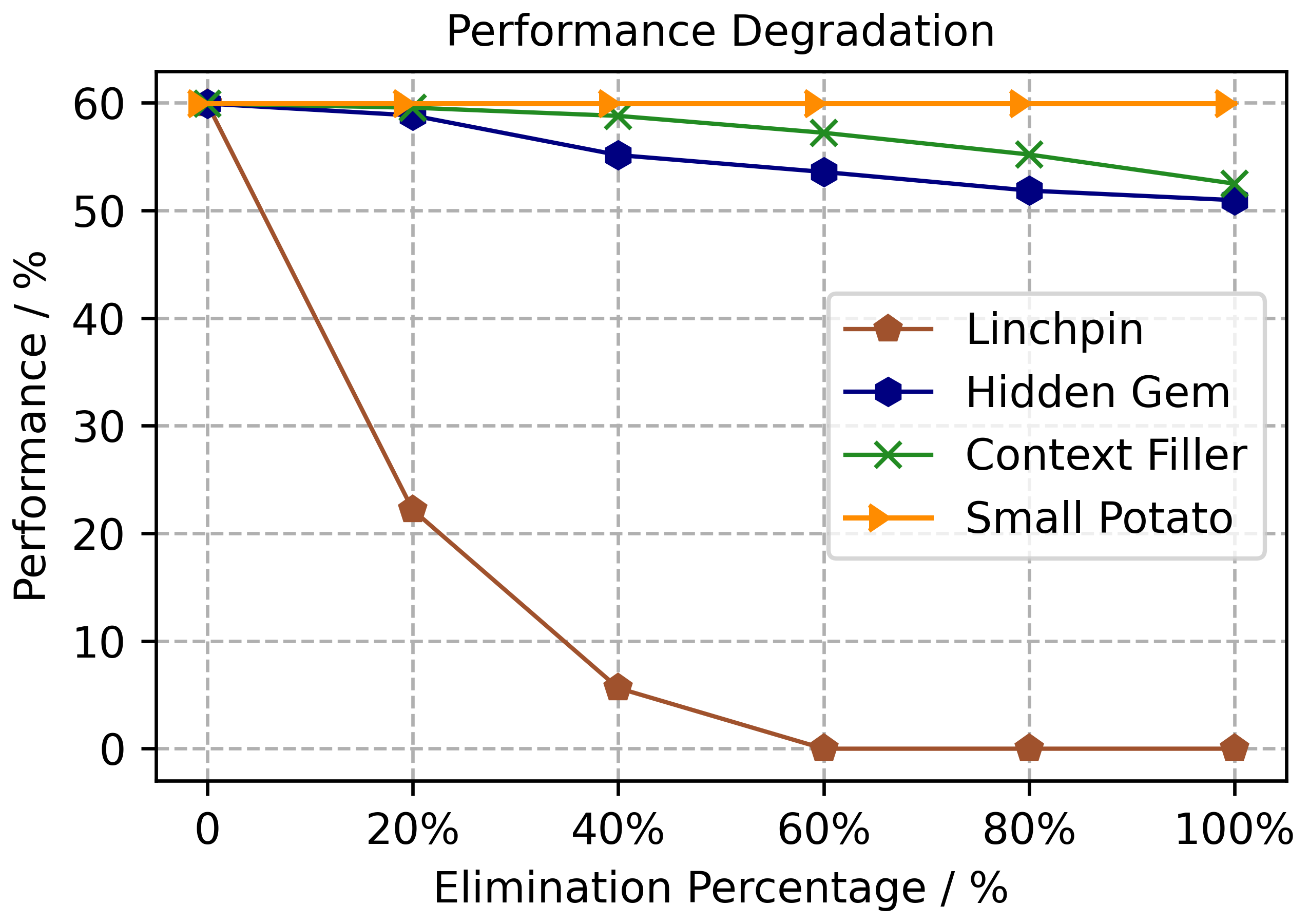}
    \caption{Performance degradation of LLaMA-2-7B-Chat after attention elimination on five LongBench tasks. Tokens with attention scores exceeding 0.05 are classified as Linchpins, those with scores in the 0.01–0.05 range (depending on their position) as Context Fillers or Hidden Gems, and those below 0.01 as Small Potatoes.}
    \label{fig:perf_drop}
\end{figure}

\section{Preliminaries}
\def\mX{\mathbf{X}}
\def\mA{\mathbf{A}}
\def\mD{\mathbf{D}}
\def\mQ{\mathbf{Q}}
\def\mK{\mathbf{K}}
\def\mV{\mathbf{V}}
\def\Wq{\mathbf{W}_q}
\def\Wk{\mathbf{W}_k}
\def\Wv{\mathbf{W}_v}
\def\R{\mathbb{R}}
\def\mP{\mathbf{P}}
\def\vf{\mathbf{f}}
\def\vx{\mathbf{x}}
\def\vk{\mathbf{k}}
\def\vq{\mathbf{q}}
\def\vu{\mathbf{u}}
\def\di{\mathrm{i}}

\textbf{Attention Mechanism} Given the input token embeddings as $ \mX \in \R^{n\times d} $, the attention mechanism in transformers can be computed as: 
\begin{equation}
\label{eq:attn}
    \mathrm{Softmax} \left( \mQ \mK^{\top}/{\sqrt{\mathrm{C}}} \right) \mV = \mD \mA \mV
\end{equation}
where $\mQ=\mX\Wq, \mK=\mX\Wk, \mV=\mX\Wv$ are Query, Key, Value matrices, $\mathrm{C}$ is a scaling factor, and $ \Wq, \Wk, \Wv \in \R^{d \times d} $ are projection matrices. Since $\mathrm{Softmax}$ can be regarded as a dynamic nonlinear scaling of KV similarity $\mA$, we can use $\mD \in \R^{d \times d}$ to integrate $\mathrm{C}$ and $\mathrm{Softmax}$ for a direct representation, where $\mD$ is dependent on $\mA$.

\textbf{Rotary Position Embedding} Transformer models require explicit positional information to be injected. We only consider RoPE~\citep{su2023roformer} here, which is frequently used in many LLMs \citep{touvron2023llama,jiang2023mistral}. Given a position index $m \in [0, c)$ and $\mX := [x_0, x_1, \ldots, x_{d}]^\top$, RoPE defines a vector-valued complex function $\vf(\mX, m)$ as follows:
\begin{eqnarray}
\label{eq:rope}
    \vf(\mX,m) &= &[(x_0 + \di x_1) e^{\di m \theta_0}, \nonumber \\
    &\ldots&, (x_{d-2} + \di x_{d-1})e^{\di m \theta_{d/2-1}}]^\top
\end{eqnarray}
where $\di := \sqrt{-1}$ is the imaginary unit and $\theta_j = 10000^{-2j/d}$. In conjunction with Eq.~\ref{eq:attn}, we can also integrate RoPE into a changing coefficient matrix $\mP \in \R^{d \times d} $ to achieve scaling determined by relative positions:
\begin{equation}
\label{eq:attn_rope}
   \mathrm{Softmax}(\mathrm{Re}\langle\vf(\mQ, m), \vf(\mV, n)\rangle) = \mD \mP \mA
\end{equation}
After this change, $\mD$ is dependent on both $\mP$ and $\mA$. 

\section{Methodology}
\def\mW{\mathbf{W}}
\def\mWa{\mathbf{W}_{A}}
\def\mWin{\mathbf{W}_{in}}
\def\mWou{\mathbf{W}_{ou}}
\def\pin{\mathrm{Pick}_{in}}
\def\pou{\mathrm{Pick}_{ou}}
\def\ein{\mathrm{E}_{in}}
\def\eou{\mathrm{E}_{ou}}

In this chapter, we first present our reasoning process and then introduce our method. We provide intuitive and easy-to-understand reasoning in the main text, with more analyses available in the appendix.

\subsection{Analysis} 
\paragraph{Tokens Play Different Roles}
Through our experiments and analyses, we \textit{first} defined that tokens in attention mechanisms can be categorized into 4 types, and their effects on performance after elimination are shown in Figure~\ref{fig:perf_drop}.

\begin{figure}
    \centering
    \includegraphics[width=0.7\linewidth,height=4.4cm]{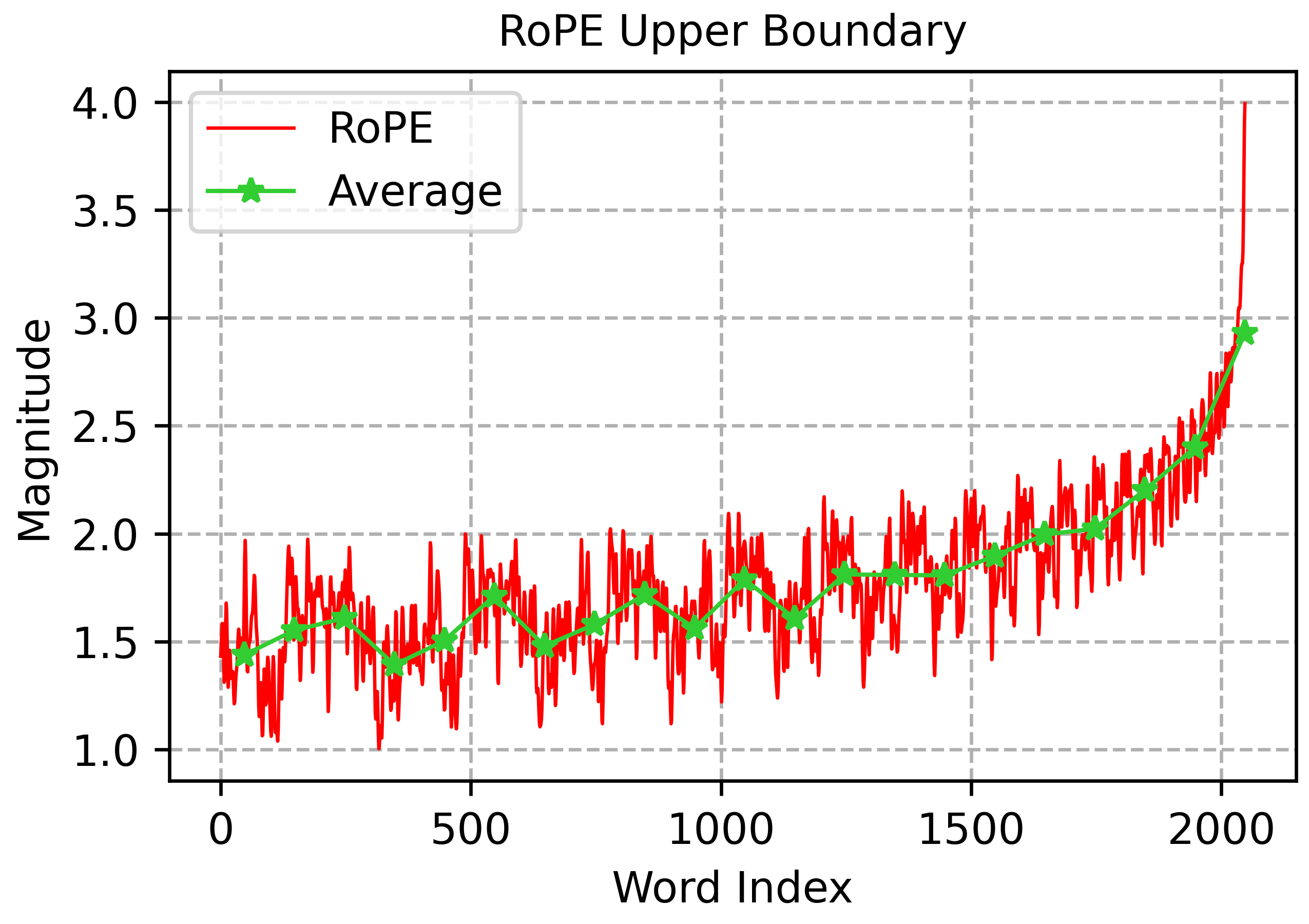}
    \caption{RoPE upper boundary alongside its averaged counterpart at intervals of 100-word index.}
    \label{fig:rot_max}
\end{figure}

\textbf{Linchpins} $\mX_{lin}$: Tokens with significantly high attention scores. These tokens often appear near the current token or the first token~\cite{xiao2023streamingllm} and frequently account for over $70\%$ of the accumulated attention scores. These tokens often have a critical impact on the model's reasoning results, as they are the primary contributors to altering hidden states between layers under the residual structure~\cite{liu2024minicache} and causing outliers~\cite{bondarenko2023quantizable}.

\textbf{Context Fillers} $\mX_{con}$: Tokens near the current token and exhibit relatively high attention scores. They generally account for approximately $25\%$ of the total accumulated attention scores but with constrained maximum value. Their presence has only a limited impact on generation results, as the model's reasoning capability is affected (not large) only when a large amount of them are eliminated.

\textbf{Hidden Gems} $\mX_{hid}$: Tokens located in distant regions yet exhibiting noticeably higher attention scores. Despite their distance from the current token, these tokens exert a more pronounced impact on performance than $\mX_{con}$.

\begin{figure}
    \centering
     \includegraphics[width=1\linewidth]{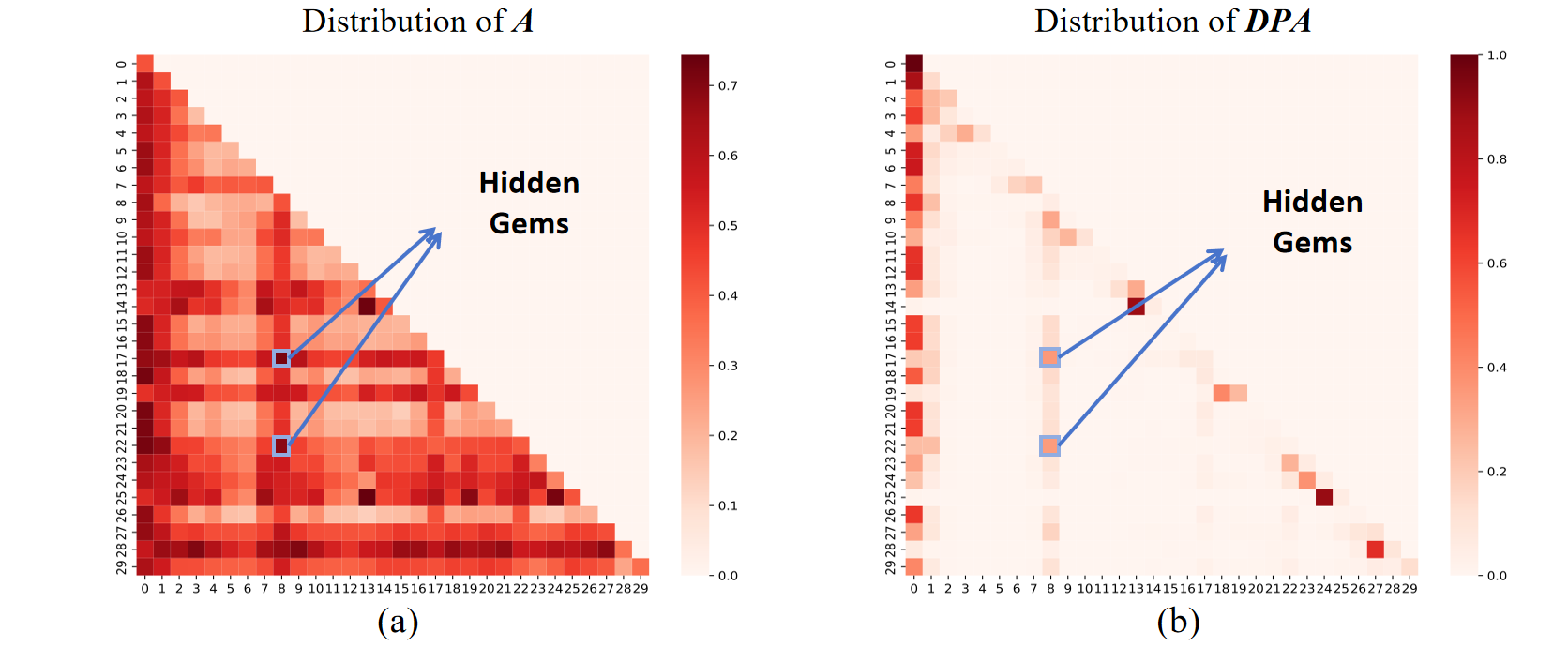}
    \caption{(a) Normalized distribution of $\mA$. (b) Normalized distribution of $\mD \mP \mA$. We aim to identify those Hidden Gems (\textcolor{blue}{blue boxes}) with high similarity at distant positions.}
    \label{fig:hidden_gem}
\end{figure}

\textbf{Small Potatoes} $\mX_{pot}$: The vast majority of tokens with sparse attention. Contribute generally nothing, with accumulated attention scores no more than $2\%$.

Intuitively, identifying $\mX_{hid}$ to enhance the model's retrieval ability is a reasonable approach. These hidden gems are expected to have high relevance with the current token, but their influence is significantly constrained due to the effects of RoPE and Softmax. Specifically, the sublinear decay ratio of RoPE at greater distances (Figure~\ref{fig:rot_max}) combined with the exponential scaling of Softmax results in $\mX_{hid}$, despite their high similarity, only barely maintaining the magnitude of attention scores as $\mX_{con}$ after the scaling of $ \mD \mP$, as shown in Figure~\ref{fig:hidden_gem}. From a mathematical perspective, leveraging the properties of RoPE and softmax, the classification of these four types of tokens corresponds to four distinct scaling behaviors of attention scores $\mA$ in both positional and magnitude spaces, as elaborated in Appendix~\ref{sec:attn_dist}.

\begin{figure*}[!t]
    \centering
     \includegraphics[width=0.9\linewidth]{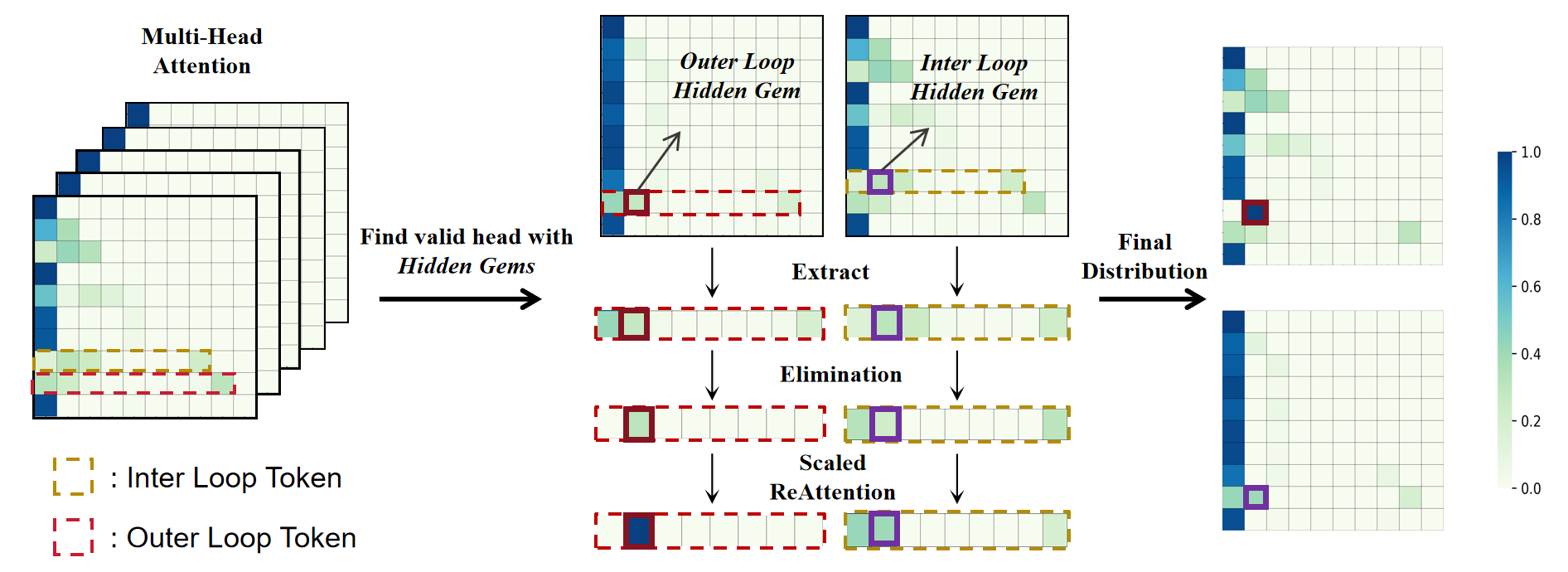}
    \caption{\textbf{Overall Pipeline.} SRA first identifies heads where the inter/outer loop contains Hidden Gems and then extracts them for attention elimination. The eliminated attention scores will be amplified (Scaled) and redistributed to these Hidden Gems (ReAttention).}
    \label{fig:pipeline}
\end{figure*}

\paragraph{Information Is Transferred Step by Step}
Under the combined effects of RoPE and softmax, attention cannot focus on tokens that are very distant from the current token, making it impossible to directly access information from distant tokens. We conducted experiments to measure how accumulated attention scores on keywords change across layers with varying distances. We found that beyond a certain distance, the accumulated attention score on keywords becomes minimal, yet the model is still able to produce normal outputs. A reasonable explanation for this is that information is continuously propagated during the inter-layer propagation process, ultimately being received by the current token. See more details in the Appendix~\ref{sec:info_trans}

\paragraph{LLMs Are Inherently Stable}
In line with our attention elimination approach and previous findings, we observed that removing $30\%$ of the accumulated attention scores across all tokens in all layers of LLMs—including most $\mX_{con}$ and $\mX_{hid}$, as well as all $\mX_{pot}$—still allowed the model to output content stably, albeit with some performance degradation on complex tasks, as shown in Figure~\ref{fig:perf_drop}. Therefore, a \textit{moderate increase in attention scores} should also not lead to large disturbances. Based on our previous analysis, by manually identifying and amplifying $\mX_{hid}$, we obtained exciting results: the model's information comprehension and retrieval abilities improved significantly for long texts! This insight was pivotal in driving the creation of SRA.

\subsection{Scaled ReAttention}
Based on all the analyses above, we designed the \textbf{S}caled \textbf{R}e\textbf{A}ttention (SRA) technique with two loops: an inter-loop and an outer-loop. The inter-loop is responsible for reinforcing the transfer of information, achieved by selecting an intermediate subset of tokens and strengthening their connection with preceding tokens. Meanwhile, the outer-loop helps the final subset of tokens ignore the distance constraints introduced by PE, allowing them to allocate attention to distant tokens directly. Both loops first identify the regions to enhance. Then, they eliminate the majority of the attention weights among the selected tokens. The erased attention weights are amplified and redistributed to those Hidden Gems within the region. The overall framework is illustrated in Figure~\ref{fig:pipeline}.

Note that the fundamental difference between our technique and previous ones lies in the fact that, after the softmax, we increase the attention sum of certain tokens—originally limited to 1—\textbf{to exceed 1 through SRA}. These additional, intentionally introduced attentions help improve the model's performance.

\paragraph{Identify Strengthened Blocks} 
Specifically, given an attention weight matrix $\mWa = \mD \mP \mA \in \R^{n \times n}$, we first divide it into blocks and apply the SRA operations only to specific blocks and regions. Due to the impact of the attention sink~\cite{xiao2023streamingllm}, we preserve the integrity of the first $C_s$ initial tokens. For the last $C_e$ tokens, we specify that they only participate in the outer loop. To strategically enhance distant hidden gems, for the remaining intermediate tokens, we divide them evenly into $l+3$ distinct blocks, where $\mathbf{C}_m^{l+3} = [C_m^1, \ldots, C_m^{l+3}]$ and $C_m^i$ is the initial token's index for the $i$th block in $\mathbf{C}_m^{l+3}$. 

The inter-loop SRA begins at layer 0 and ends at the penultimate layer, while the outer-loop SRA starts from the second layer and also ends at the penultimate layer. For every layer, both of them select only one block each. Given the selection algorithms $\pin$ and $\pou$ for inter-loop and outer-loop respectively, the regions of attention weights selected for the $i$th layer (starting at $0$th) are as follows:
\begin{align}
\label{eq:pick_alg}
    \mWa[\pin(\mWa, i)] &= \mWa^{C_m^{i+4}:C_m^{i+5} , C_m^{i+1}:C_m^{i+3}} \nonumber\\
    \mWa[\pou(\mWa, i)] &= \mWa^{-C_e: , C_m^{i+3}:C_m^{i+4}}
\end{align}
The indexing rules here are consistent with the indexing rules of $torch.tensor$ in \textbf{PyTorch}. This hierarchical approach ensures that the inter loop focuses on refining intermediate regions, while the outer loop further consolidates and enhances these refined regions in the subsequent layer.

\paragraph{Attention Elimination and Scaled Redistribution} 
The goal of elimination is to remove the smaller Context Fillers and Small Potatoes among the enhanced tokens while preserving the Hidden Gems as much as possible. Specifically, for the $j$th enhanced block $\mWin = \mWa^{C_m^{j}:C_m^{j+1} , :}$ for inter loop, $\mWou = \mWa^{- C_e:, :}$ for outer loop, the inter-loop eliminator $\ein$ and outer-loop eliminator $\eou$ is defined as:
\begin{align}
\label{eq:elim_alg}
    \ein(\mWin, j) &= \mathrm{Whe}(\mWin > (\tau_{in} / C_m^{j}), \mWin, 0) \nonumber\\
    \eou(\mWou) &= \mathrm{Whe}(\mWou > (\tau_{ou}/C_r), \mWou, 0)
\end{align}
Here, $\mathrm{Whe}$ functions the same as $torch.where$ and $C_r = n - C_e$. If no Hidden Gems are found during the elimination process, such as all $\mWin^{:,C_m^{j-3}:C_m^{j-2}}=0$, the elimination will be skipped, and no subsequent operations will be performed. Otherwise, the erased weights will be summed in a token-wise manner and multiplied by a scaling factor, $s_{in}$ for inter loop and $s_{ou}$ for the outer loop. This amplification enhances performance by sacrificing the stability of the LLM, allowing the accumulated attention to exceed 1. Finally, the amplified erased weights will be evenly re-added on those uneliminated Hidden Gems within targeted blocks in Eq.~\ref{eq:pick_alg}. The inter-loop algorithm is exhibited in Algorithm~\ref{algo:sra_inter}, while the outer-loop one is in the Appendix~\ref{sec:outler_sra}. During inference, SRA is triggered \textit{only in the prefilling stage}.

\begin{algorithm}[h]
   \caption{Inter-loop Scaled ReAttention}
\begin{algorithmic}
    \label{algo:sra_inter}
   \STATE {\bfseries After applying Softmax on attention weights:}
   \STATE {\bfseries Input:} \textit{Attention Weights} $\mWa$,  \textit{Layer Index} $i$,  \textit{Layer Num} $l$,  \textit{Inter Threshold} $\tau_{in}$,  \textit{Inter Scaling Factor} $s_{in}$. \COMMENT{\textbf{Note}: All unspecified functions are from \textbf{PyTorch}.}
   \IF[Inter Loop]{$ (l-1) > i > 0 $}
    \STATE /* Function only on indexes having Hidden Gems */
    
    \STATE $\mathbf{idx}_{gem} = any(\mWa[\pin(\mWa,i)] > (\tau_{in} / C_m^{i+4}))$ 
    
    \STATE $\mathbf{\mW}_{eli} = \ein(\mWa[\mathbf{idx}_{gem}], i+4)$ 
    
    \STATE $\mathbf{idx}_{tar} = \pin(\mathbf{\mW}_{eli},i)$
    
    \STATE $\mW_{tar} = \mW_{eli}[\mathbf{idx}_{tar}]$
    
    \STATE /* Prepare scaled attention removal */
    \STATE $\mW_{re} = sum(\mW_{eli},dim=-1)$
    
    \STATE $\mW_{rm} = oneslike(\mW_{re})-\mW_{re}$
    \STATE $\mathbf{m}_{gem} = where(\mW_{tar} > 0, 1, 0.01)$
    
    \STATE $\mW_{add} = div(\mW_{rm}, sum(\mathbf{m}_{gem}, dim=-1))*s_{in} $
    \STATE /* Readded to original weights */
    
    \STATE $\mW_{eli}[\mathbf{idx}_{tar}] = \mW_{tar} + \mW_{add}$
    \STATE $\mWa[\mathbf{idx}_{gem}] = \mW_{eli}$
    \ENDIF 
\end{algorithmic}
\end{algorithm}

\section{Experiments}

\subsection{Setting}
Our experiments comprehensively demonstrate the effectiveness of our method from multiple perspectives. We selected commonly used model series such as LLaMA~\cite{touvron2023llama}, Mistral~\cite{jiang2023mistral}, Qwen~\cite{bai2023qwen}, and LongChat~\cite{longchat2023}, as well as their YaRN~\cite{peng2023yarn} and LandMark~\cite{mohtashami2023landmark} variants, as baseline models. First, we used methods from LandMark~\cite{mohtashami2023landmark} and LongChat~\cite{longchat2023} to evaluate the improvement in retrieval capabilities brought by SRA. Next, we tested the model's ability to summarize and understand long, complex texts on the XSUM~\cite{narayan2018xsum} dataset under GPT-4 evaluation protocol~\cite{vicuna}. We further validated the superiority of our approach through downstream tasks on publicly available long-text comprehension benchmarks, including LongBench~\cite{bai2023longbench}, LongBench v2~\cite{bai2024longbench}, InfiniteBench~\cite{zhang2024bench}. 

The configuration of SRA is not a one-size-fits-all solution. Instead, it requires dynamic tuning based on the requirements of specific tasks. Several factors influence the choice of SRA parameters, including task characteristics and variations among different baselines. In our experiments, we typically keep the total accumulated attention of SRA-strengthened tokens within the range of 1.1 to 1.4. Detailed discussions can be found in the Appendix~\ref{sec:sra_config}.

\subsection{Reterieval Evaluation}
\label{sec:retrieval_eval}
\begin{figure*}[t]
    \centering
     \includegraphics[width=1\linewidth]{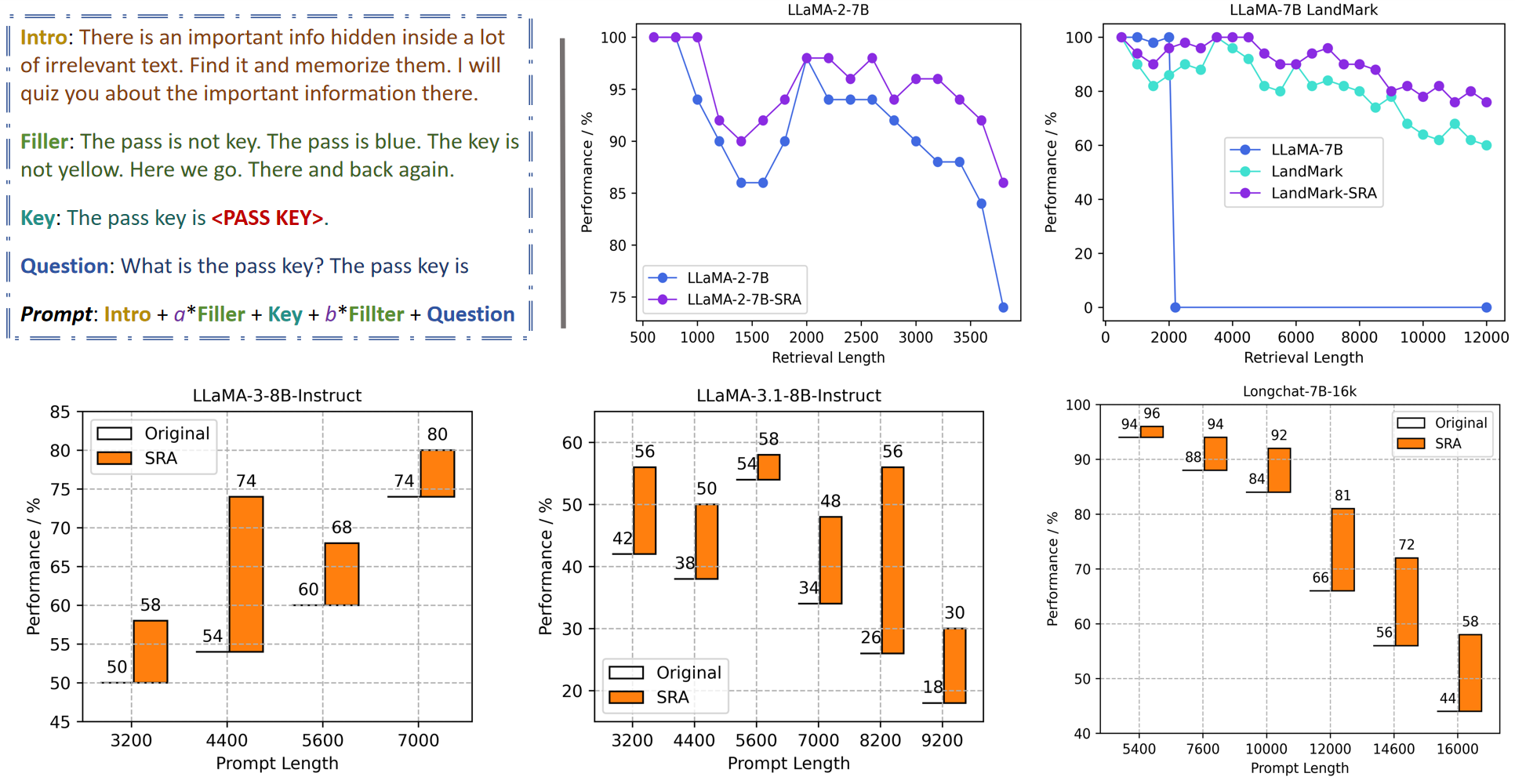}
    \caption{Retrieval Results. (\textbf{Top Left}) Modified Landmark retrieval prompt. Here, a and b are scaling factors used to adjust the retrieval distance. (\textbf{Top Right}) The retrieval results on LandMark. (\textbf{Bottom}) The retrieval results on LongChat. Our results indicate that SRA substantially improves retrieval capabilities across various models and tasks, highlighting its versatility and effectiveness.}
    \label{fig:retrieval_perf}
\end{figure*}

We began by assessing the improvements in retrieval capabilities introduced by SRA within the LongChat framework, followed by an evaluation using a retrieval prompt proposed in LandMark. We modified the original retrieval prompt to increase complexity. For the \textit{PASS KEY}, we randomly generated 50 words comprising numbers and uncommon vocabulary. By varying the retrieval distance, we tested the model’s performance. Throughout the experiments, \(a\) was fixed at 8, while \(b\) was varied at intervals of 200 tokens. The prompt and results of the two tasks are shown in Figure~\ref{fig:retrieval_perf}. 

Experiments reveal a significant enhancement in the model's retrieval capabilities after incorporating SRA. For the LandMark \textit{PASS KEY} retrieval task, \textbf{LLaMA-2-7B} achieves an average improvement of 4.7\% over the original model. Additionally, compared to the LandMark variant of \textbf{LLaMA-7B}, our approach delivers an average improvement of 8.5\%, effectively enabling SRA to mitigate the decline in retrieval performance caused by its structure-altering. On the LongChat benchmark, SRA achieves a notable performance boost, with an average retrieval accuracy improvement exceeding 10\% over these original models.

\begin{table*}[tb]
\footnotesize
\centering
\caption{ \textbf{LongBench Results.} We present the results of three open-source models evaluated on seven tasks from LongBench, both before and after applying SRA. SRA delivers consistent performance improvements without requiring any additional fine-tuning or retraining.}
{
\begin{tabular}{l|ccccccc|c}
\toprule
\textbf{Model / Tasks}$\uparrow$  & \textbf{MFQA-EN} & \textbf{VCSUM}  & \textbf{TREC} & \textbf{SAMSum} & \textbf{LSHT} & \textbf{LCC} & \textbf{RepoBench-P} & \textbf{Average}\\ \hline

LLaMA-2-7B-Chat & 36.22 & 15 & 64.5 & 40.7 & 17.75 & 58.50 & 52.45 & 40.73\\ 

\cellcolor{Gray}\textbf{SRA} & \cellcolor{Gray}37.83 & \cellcolor{Gray}21.0 & \cellcolor{Gray}66.5 & \cellcolor{Gray}42.31 & \cellcolor{Gray}19.00 & \cellcolor{Gray}59.36 & \cellcolor{Gray}53.23 & \cellcolor{Gray}42.74(+2.01) \\

\midrule

LLaMA-3-8B-Instruct & 41.50 & 14.8 & 75.5 & 42.48 & 24.25 & 58.87 & 50.73 & 44.01\\ 

\cellcolor{Gray}\textbf{SRA} & \cellcolor{Gray}42.71 & \cellcolor{Gray}17.5 & \cellcolor{Gray}78.5 & \cellcolor{Gray}43.04 & \cellcolor{Gray}28.00 & \cellcolor{Gray}59.72 & \cellcolor{Gray}51.27 & \cellcolor{Gray}45.82(+1.81) \\

\midrule

LongChat-v1.5-7B-32k & 41.40 & 9.9 & 63.5 & 34.20 & 23.20 & 53.00 & 55.30 & 40.07\\ 

\cellcolor{Gray}\textbf{SRA} & \cellcolor{Gray}43.20 & \cellcolor{Gray}14.5 & \cellcolor{Gray}65.1 & \cellcolor{Gray}35.80 & \cellcolor{Gray}25.10 & \cellcolor{Gray}53.72 & \cellcolor{Gray}55.96 & \cellcolor{Gray}41.91(+1.84) \\

\bottomrule
\end{tabular}
}
\label{tab:longbench_v1}
\end{table*}

\subsection{Summarization Evaluation}
\label{sec:xsum}
\begin{figure}
    \centering
     \includegraphics[width=1\linewidth]{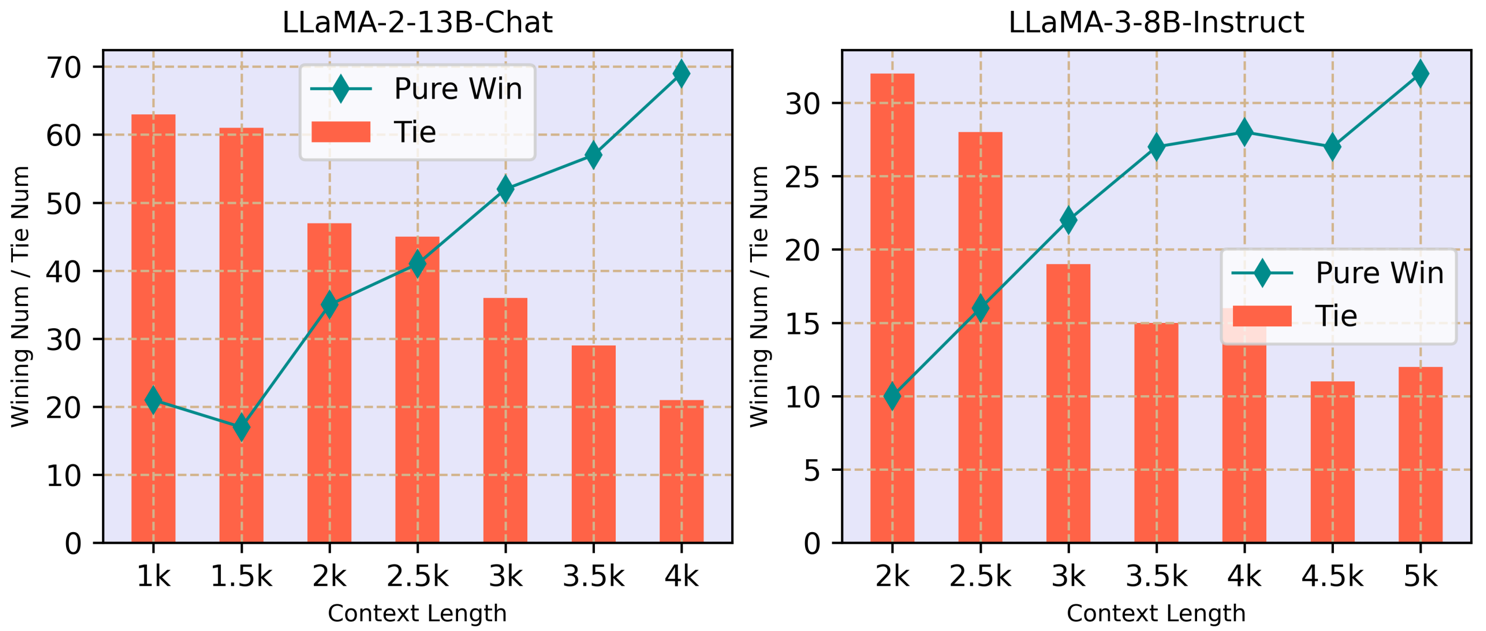}
    \caption{\textbf{Xsum results on LLaMA-2-13B-Chat and LLaMA-3-8B-Instruct}. SRA has demonstrated significant advantages in long-text comprehension and summarization, with these benefits becoming increasingly pronounced as the text length grows.}
    \label{fig:xsum_perf}
\end{figure}

We tested the enhancements brought by SRA on texts of different lengths using \textbf{LLaMA-3-8B-Instruct} and \textbf{LLaMA-2-13B-Chat}. Specifically, for \textbf{LLaMA-3-13B-Chat}, we started with texts of length 1000 tokens, collecting 100 cases at intervals of 500 tokens, up to a length of 4000 tokens. For \textbf{LLaMA-3-8B-Instruct}, we started with texts of length 2000 tokens, collecting 50 cases at intervals of 500 tokens, up to a length of 5000 tokens. We used GPT4o as the evaluation model following the GPT-4 evaluation protocol, comparing the outputs under SRA with the original model outputs. The results are illustrated in Figure~\ref{fig:xsum_perf}, where we show the counts of ``pure win" and ``tie" cases. Here, the ``pure win" refers to the winning number of SRA minus the winning number of the original model. 

The results indicate that the benefits of SRA become increasingly evident as text length grows, with a declining number of ties and a steadily rising count of ``pure wins". Beyond a context length of 3000 for \textbf{LLaMA-2-13B-Chat} and 3500 for \textbf{LLaMA-3-8B-Instruct}, over half of the total samples show improvements compared to the original results when SRA is applied.

\subsection{Results on Open-Source Benchmarks}
Starting with LongBench, we selected 7 tasks including MultiFieldQA-EN (MFQA-EN), VCSUM~\cite{wu-etal-2023-vcsum}, TREC~\cite{li2002learning}, SAMSum~\cite{gliwa2019samsum}, LSHT~\cite{LSHT}, LCC~\cite{guo2023longcoder}, and RepoBench-P~\cite{liu2024lost}. The following results are illustrated in Table~\ref{tab:longbench_v1}. Our SRA technique enables an overall performance gain exceeding 1.8 across all models.

We further explored the performance of SRA on \textbf{YaRN-Mistral} within the InfiniteBench benchmark, with the results shown in Table~\ref{tab:infini_results}. With the enhancements provided by SRA, the decline in retrieval and comprehension capabilities caused by PI modifications introduced by YaRN was significantly mitigated. This improvement further unlocks the potential of methods like YaRN and other PI approaches in the application of LLMs.

\begin{table}[tb]
    \centering
    \footnotesize
    \caption{\textbf{InfiniteBench Results of YaRN-Mistral}. Here, \textit{Retrieve} encompasses both \textit{Retrieve.PassKey} and \textit{Retrieve.Number}. SRA notably enhances the retrieval capabilities of models trained with YaRN.}
    \begin{tabular}{lcccc}
    \toprule
    ~ & Retrieve & En.Sum & En.MC \\
    \hline
    YaRN-Mistral & 74.66 & 9.09 & 27.95 \\
    \midrule
    \cellcolor{Gray}\textbf{SRA} & \cellcolor{Gray}81.24 & \cellcolor{Gray}12.08 & \cellcolor{Gray}37.25 \\
    \bottomrule
    \end{tabular}
    \label{tab:infini_results}
\end{table}

Finally, we conducted tests on the newly released LongBench v2, which includes a series of models using RoPE as their positional encoding method. The results are presented in Table~\ref{tab:longbench_v2}. The results demonstrate that SRA exhibits excellent generalizability for LLMs utilizing RoPE as their positional encoding, significantly enhancing comprehension capabilities while maintaining high compatibility with CoT prompt engineering. For smaller models, the improvements brought by SRA are particularly pronounced, with most tasks on \textbf{Llama-3.1-8B-Instruct} achieving gains of over 2\%. For larger LLMs, the most notable improvements are observed in long-text processing, with performance increases exceeding 2\% in certain tasks. This is likely because advanced LLMs are already well-optimized for handling shorter contexts effectively. Moreover, regardless of task difficulty, the enhancements achieved through SRA remain consistent, demonstrating the robustness and reliability of our method.


\begin{table*}[t]
\centering
\caption{\textbf{SRA evaluation results (\%) on LongBench v2.} Results under \colorbox{Gray}{CoT} prompting are highlighted with a gray background. SRA exhibits robust compatibility and enhancement effects for models employing RoPE as the positional encoding method, especially when integrated with CoT reasoning.}
\resizebox{\linewidth}{!}{
\begin{tabular}{p{5cm}|m{0.7cm}m{0.7cm}|m{0.7cm}m{0.7cm}|m{0.7cm}m{0.7cm}|m{0.7cm}m{0.7cm}|m{0.7cm}m{0.7cm}|m{0.7cm}m{0.7cm}}
\toprule
 &  & & \multicolumn{4}{|c}{\textbf{Difficulty}} & \multicolumn{6}{|c}{\textbf{Length ($<$32k; 32k-128k; $>$128k)}} \\
\cmidrule(r){1-3} \cmidrule(lr){4-7} \cmidrule(l){8-13}
\textbf{Model} & \multicolumn{2}{c|}{\textbf{Overall}} & \multicolumn{2}{c|}{\textbf{Easy}} & \multicolumn{2}{c|}{\textbf{Hard}} & \multicolumn{2}{c|}{\textbf{Short}} & \multicolumn{2}{c|}{\textbf{Medium}} & \multicolumn{2}{c}{\textbf{Long}} \\ 
\midrule

{Llama-3.1-8B-Instruct} 
& 30.0 & \cellcolor{Gray}30.4 
& 30.7 & \cellcolor{Gray}36.5 
& 29.6 & \cellcolor{Gray}26.7 
& 35.0 & \cellcolor{Gray}34.4 
& 27.9 & \cellcolor{Gray}31.6 
& 25.9 & \cellcolor{Gray}21.3 \\ 

\textbf{SRA} 
& 31.3 & \cellcolor{Gray}32.0 
& 31.9 & \cellcolor{Gray}38.2 
& 30.7 & \cellcolor{Gray}28.5 
& 37.3 & \cellcolor{Gray}36.9 
& 29.3 & \cellcolor{Gray}33.1 
& 27.2 & \cellcolor{Gray}23.4 \\

\midrule

{Llama-3.3-70B-Instruct} & 29.8 & \cellcolor{Gray}36.2 & 34.4 & \cellcolor{Gray}38.0 & 27.0 & \cellcolor{Gray}35.0 & 36.7 & \cellcolor{Gray}45.0 & 27.0 & \cellcolor{Gray}33.0 & 24.1 & \cellcolor{Gray}27.8 \\

\textbf{SRA} 
& 31.0 & \cellcolor{Gray}37.2 
& 35.1 & \cellcolor{Gray}39.2 
& 27.9 & \cellcolor{Gray}35.8 
& 37.5 & \cellcolor{Gray}48.1 
& 28.6 & \cellcolor{Gray}34.3 
& 25.8 & \cellcolor{Gray}28.4 \\

\midrule

{Qwen2.5-72B-Instruct} 
& 39.4 & \cellcolor{Gray}38.8 
& 43.8 & \cellcolor{Gray}42.2 
& 36.7 & \cellcolor{Gray}36.7 
& 44.4 & \cellcolor{Gray}50.0 
& 34.0 & \cellcolor{Gray}28.8 
& 41.7 & \cellcolor{Gray}39.8 \\

\textbf{SRA} 
& 41.2 & \cellcolor{Gray}41.4 
& 44.5 & \cellcolor{Gray}43.6 
& 36.9 & \cellcolor{Gray}39.1 
& 45.2 & \cellcolor{Gray}51.3 
& 35.9 & \cellcolor{Gray}31.4 
& 43.5 & \cellcolor{Gray}41.6 \\

\midrule

{Mistral-Large-Instruct-2407} & 26.6 & \cellcolor{Gray}33.6 & 29.7 & \cellcolor{Gray}34.4 & 24.8 & \cellcolor{Gray}33.1 & 37.8 & \cellcolor{Gray}41.1 & 19.5 & \cellcolor{Gray}31.2 & 22.2 & \cellcolor{Gray}25.9 \\

\textbf{SRA} 
& 28.0 & \cellcolor{Gray}34.6 
& 30.5 & \cellcolor{Gray}36.2 
& 26.3 & \cellcolor{Gray}34.5 
& 39.1 & \cellcolor{Gray}43.4 
& 20.2 & \cellcolor{Gray}31.5 
& 24.3 & \cellcolor{Gray}27.2 \\

\bottomrule
\end{tabular}
}
\label{tab:longbench_v2}
\end{table*}

\subsection{Ablation Studies}
\label{sec:ablation}
In SRA, both the inter loop and outer loop are integral to its effectiveness. Even without scaling—where \( s_{in} \) and \( s_{ou} \) are set to \( 1 \)—the basic ``ReAttention" mechanism reduces perplexity during inference, as demonstrated in Table~\ref{tab:ablation_ppl}. Furthermore, extensive experiments reveal that the inter loop and outer loop enhance distinct aspects of model comprehension, as illustrated in Table~\ref{tab:ablation_tasks}. The inter loop primarily bolsters overall comprehension, making it particularly effective for tasks such as dialogue, summarization, and document understanding. In contrast, the outer loop excels in improving retrieval capabilities, especially for questions or keywords positioned near the end of prompts. Combining all of the components, SRA finally renders its superior effects.

\begin{table}[tb]
    \centering
    \footnotesize
    \caption{\textbf{Ablation study of LLaMA-3-8B on WikiText.} We use the last 200 words to calculate the perplexity.}
    \begin{tabular}{p{2cm}|m{1cm}|m{1cm}|m{1cm}}
    \toprule
    Word Counts & 1024 & 1536 & 2048 \\
    \midrule
    Original & 5.58 & 5.61 & 5.22  \\
    +inter loop & 5.57 & 5.60 & 5.21 \\
    +outer loop & 5.57 & 5.59 & 5.20 \\
    \midrule
    RA & 5.56  & 5.57 & 5.19 \\
    \bottomrule
    \end{tabular}
    \label{tab:ablation_ppl}
\end{table}

\begin{table}[tb]
    \centering
    \footnotesize
    \caption{\textbf{Ablation study of LLaMA-3-8B-Instruct on downstream tasks.}}
    \begin{tabular}{p{2.3cm}|c|c|c}
    \midrule
    Tasks & LongChat & MFQA-EN & LSHT \\
    \midrule
    Original & 59.5 & 41.50 & 24.25\\
    +inter loop & 62.5 & 42.35 & 26.80 \\
    +outer loop & 68.9 & 41.78 & 25.20 \\
    \midrule
    SRA & 70.0 & 42.71 & 28.00\\
    \bottomrule
    \end{tabular}
    \label{tab:ablation_tasks}
\end{table}

\subsection{Discussion of SRA Configurations}
In our experiments, we tested numerous sets of SRA parameters to validate the gains brought by SRA. Despite variations in models and tasks, we derived a generalizable approach for tuning SRA parameters. More discussions and their corresponding experiments are exhibited in the Appendix~\ref{sec:sra_config}.

From a model perspective, training methods and following tasks influence a model's sensitivity to SRA. For example, while both are based on LLaMA, the LongChat series demonstrates greater sensitivity to SRA compared to the LLaMA series. Generally, models trained for retrieval tasks require a lower elimination threshold and smaller scale factors. Excessive values for these parameters can lead to incorrect outputs even when the correct position is identified, such as retrieving the correct context but returning an incorrect number in retrieval tasks. For models trained under standard conditions, larger parameter values are needed, particularly for the scaling factor, which directly affects the enhancement achieved.

In terms of context length, longer texts generally require smaller scaling factors. This is because the robustness of LLMs is limited, and excessively large scaling factors can impair the model's language capabilities. Specifically, this manifests as fragmented sentences and incoherent expressions. While some keywords may still appear, they fail to form continuous and meaningful statements.

From a task perspective, as discussed in Sec.~\ref{sec:ablation}, the type of task significantly influences the configuration of the inter loop and outer loop scaling factors. For tasks such as QA and summarization, larger \( s_{in} \) and smaller \( s_{ou} \) are recommended. In contrast, for retrieval-based tasks, focusing the gains from SRA on the keywords in the final question—by setting a larger \( s_{ou} \)—yields better results.

\section{Limitations}
\paragraph{Variability of SRA}
Although we have established a relatively general set of guidelines, a small amount of task-specific testing remains unavoidable. This is particularly important for adjusting SRA parameters to suit different tasks. However, our experiments reveal that models within the same series generally exhibit similar characteristics, reducing the need for extensive testing to some extent. 

Excessive application of SRA can cause significant disruptions to LLMs, potentially rendering them non-functional. Under normal use, SRA is characterized by a slight increase in perplexity compared to the original model. While some negative effects are present, the positive outcomes far outweigh them. From a task perspective, this slight increase in perplexity under normal usage has no impact on the quality or accuracy of the generated content.

\paragraph{Inference Efficiency}
A notable limitation of SRA is its reliance on explicit manipulation of attention scores after Softmax, as operations before the Softmax stage would disrupt the original distribution and introduce significant interference. This explicit computation prevents the use of certain attention acceleration techniques, such as FlashAttention~\cite{dao2022flashattention}, in conjunction with SRA, leading to slower inference speeds. However, the impact of SRA's operations on pure processing time is minimal, as we only enhance a small fraction of tokens in heads with Hidden Gems and just in the prefilling stage. Comparisons are shown in Table~\ref{tab:inference_speed}. Moreover, the performance gains provided by SRA without additional training fully compensate for the impact of the extra time.

\begin{table}[tb]
    \centering
    \footnotesize
    \caption{\textbf{Inference Speed of LLaMA 7B.} We report the running memory (denoted as `RM') and speed in NVIDIA A100-80G.}
    \begin{tabular}{p{2cm}|p{1.5cm}|p{1.5cm}}
    \toprule
    Method & RM & Token/s \\
    \midrule
    Normal & 14.4 G &  69.2 \\
    Flash Attention& 13.7 G &  77.9  \\
    \midrule
    SRA & 14.6 G &  64.8  \\
    \bottomrule
    \end{tabular}
    \label{tab:inference_speed}
\end{table}

\section{Conclusion}
In this paper, we introduce SRA, a training-free method designed to enhance the contextual understanding capabilities of large language models. SRA achieves this by manually adjusting attention scores, amplifying the scores projected onto Hidden Gems, and trading off some model stability to improve retrieval and comprehension abilities. Through extensive experiments, we demonstrate the effectiveness of SRA across a variety of tasks, achieving significant performance improvements in retrieval and summarization tasks. Furthermore, SRA delivers notable enhancements even in open-ended long-text scenarios.

\section*{Impact Statement}
Our goal is to enhance large language models' reading comprehension and information retrieval capabilities without requiring any additional training. Our research is strongly oriented toward the industry, where cost is a crucial factor. Unlike previous research-focused work that requires significant resource investment to boost performance, our study emphasizes lightweight industrial implementation and practical deployment. In our view, our research makes an outstanding contribution.

The application scenarios for our research are highly extensive, as most large language models today are based on RoPE for positional encoding. Through a series of experiments, we have demonstrated the universality of our method. For instance, it can be applied to everyday tasks such as document summarization, inductive reasoning, long-text keyword retrieval, and memory in long and complex conversations, covering nearly all daily scenarios that require handling long texts.

For this work, the key point we need to emphasize remains the same: \textbf{no additional training is required}. Compared to the hundreds or thousands of A100 hours typically needed for training or fine-tuning, achieving immediate performance improvement through a plug-and-play method is exceptionally valuable.

\nocite{langley00}

\bibliography{example_paper}
\bibliographystyle{icml2025}

\end{document}